\documentclass{article} 
\usepackage{float}
\usepackage{ijcai18}

\usepackage{times}
\usepackage{xcolor}
\usepackage{soul}
\usepackage[utf8]{inputenc}
\usepackage[small]{caption}

\usepackage{hyperref}
\usepackage{graphicx}
\usepackage{url}
\usepackage{amsfonts}
\usepackage{amssymb,amsmath}
\usepackage{cleveref}
\usepackage{dblfloatfix}
\crefname{equation}{Eq.}{Eqs.}

\graphicspath{{media/}}

\title{Deep Net Triage:\\Analyzing the Importance of Network Layers via Structural Compression}

\author{
Theodore S. Nowak$^1$, 
Jason J. Corso$^2$, 
\\ 
Robotics Institute$^{1,2}$ \\
Department of Electrical Engineering and Computer Science $^2$\\
University of Michigan \\
Ann Arbor, MI 48109, USA \\ 
tsnowak@umich.edu,
jjcorso@umich.edu
}

\begin{document}

\maketitle

\begin{abstract}
Despite their prevalence, deep networks are poorly understood. This is due, at least in part, to their highly parameterized nature. As such, while certain structures have been found to work better than others, the significance of a model's unique structure, or the importance of a given layer, and how these translate to overall accuracy, remains unclear. 
In this paper, we analyze these properties of deep neural networks via a process we term \emph{deep net triage}. Like medical triage---the assessment of the importance of various wounds---we assess the importance of layers in a neural network, or as we call it, their \emph{criticality}.
We do this by applying \emph{structural compression}, whereby we reduce a block of layers to a single layer. After compressing a set of layers, we apply a combination of initialization and training schemes, and look at network accuracy, convergence, and the layer's learned filters to assess the criticality of the layer. We apply this analysis across four data sets of varying complexity.
We find that the accuracy of the model does not depend on which layer was compressed; that accuracy can be recovered or exceeded after compression by fine-tuning across the entire model; and, lastly, that Knowledge Distillation can be used to hasten convergence of a compressed network, but constrains the accuracy attainable to that of the base model. 

\end{abstract}

\section{Introduction}

As computational devices and methods become more powerful, deep learning models are able to grow ever deeper, learn ever more complex features, and be applied to ever more problem spaces \cite{vgg16}. To grow so deep without err, the most modern of networks have relied on clever intermediary layers---such as shortcut connection layers in \cite{resnet}.
While these methods allow for learning representations afforded only by very deep architectures, it is known that there are extraneous features and excess parameters existent in these over-parameterized models \cite{OBD}. Strategies to optimize the size of these networks have been pursued along a number of routes.

The question of how to best prune redundant parameters has been the focus of deep compression methods \cite{caruana-modelcomp,dally-pruningtraininghuffman}. Others have investigated transferring the features or "knowledge" of an ensemble of models or of a single, larger, "parent" model into a smaller model via Knowledge Distillation \cite{knowledge_distillation,fitnets}. And, some have set out to find methods of generating optimal networks often via genetic algorithms \cite{genetic}.

While called structural compression, we do not use existing network compression methods, such as \cite{dally-pruningtraininghuffman}, to compress our network. We instead focus on altering the structure of the network, and assessing the importance of layers. We do, however, utilize existing Knowledge Distillation practices to help perform our triage \cite{fitnets}. 

Another approach towards optimizing the parameterization of neural networks has been to simply improve our understanding of how and what they learn. Optimization researchers have begun to study the properties of the learning space in which deep neural networks operate, and search for bounds and guarantees therein \cite{vidal}. Alternatively, culminating in the popularity of Google's Deep Dream, others have focused on understanding neural networks via the visualization and study of their filters \cite{deepdream,deepvis}.

In this paper we take the notion of compression and apply it structurally to the network model itself. We utilize Knowledge Distillation as one method of probing the network. Upon compression, though, we seek to understand how these compressive modifications have altered the model. We do so by first assessing accuracy and rate of convergence. We then verify these findings by visualizing the filters learned by the model.
Amongst existing methods, our work is most similar to that of \cite{visandunderstand}. Like Zeiler and Fergus we alter a neural network and study the effects of these alterations. Unlike this previous work, we systematically assess layers in a sequential and compressive manner in order to understand their importance. We additionally, go beyond the work of Zeiler and Fergus by analyzing the effects of this process across data sets of increasing complexity, and by concisely comparing their accuracies and convergence rates. We then verify these results with qualitative interpretation of filter activations.

We center our analysis around the VGG16 network as its structure lends itself to discretization, and thus this lens of triage and structural compression \cite{vgg16}. That is, we utilize the inherent structure of VGG16---five blocks of two to three convolution layers, followed by a max pooling layer---to frame our analysis.

We perform our analysis using a combination of initialization and training schemes. The initialization schemes used include the random initialization of weights, the initialization of weights as a combination of prior, parent filter weights, and by training a Student-Teacher network to an intermediate checkpoint. As for training schemes, we either train only on the compressed layer or across the entire network after initialization.

We apply each combination of these schemes to four data sets of increasing granularity and complexity: MNIST, CIFAR10, CIFAR100, and Standford Dogs. We seek to assess whether differences in layer criticality emerge on data sets of different complexity.

The results of our analysis are as follows. First, our experiments show that no layer is more critical. That is, for each data set maximal attainable accuracy is roughly unchanged by which layer is compressed. Secondly, that fine-tuning over the entire compressed model leads to novel, more optimal filter representations to be learned. Which in turn leads to above baseline network performance. Additionally, we show that applying a Student-Teacher framework with intermediate targets can reduce the epochs needed for a network to converge to its maximal accuracy, which lends to the notion that they can be used to train large networks more quickly. But this comes at a cost, as we find that the student network cannot learn more optimal representations than the teacher.

\section{Methods of Analysis} \label{sec:2}

Here, we first describe the concept of deep net triage. We then look at how structural compression is performed, and how it is used to compress a series of layers in a network down to a single layer. 

\subsection{Deep Net Triage}
To understand the criticality of various layers in a neural network, we seek to isolate them, and compare them amongst themselves and the original model in a consistent fashion. We do so by consecutively compressing, initializing, and training the network variants to convergence. We call this systematic manner of analysis deep net triage to liken the compression to illness and the layers of the model to systems of the body. As such we hope to evaluate which are most critical to network health and function.

\begin{figure}[t]
\centering
\includegraphics[width=.98\linewidth]{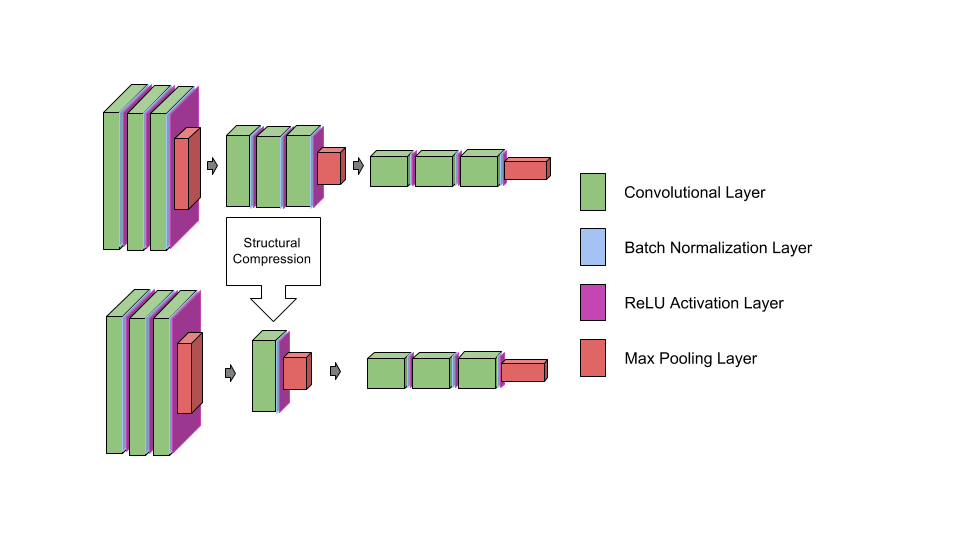}
\caption{A pictorial representation of structural compression. A series of convolutional layers is approximated by a single layer.}
\label{fig:struct_comp}
\end{figure}

\subsection{Structural Compression}

The VGG16 network is comprised of five blocks of convolutional layers, each of which are separated by max pooling layers. Within each block, the number of convolutional filters per layer is held constant. To perform structural compression, we take one such block and approximate the functions learned by the two or three layers therein with a single layer, $f_c$. This is also depicted pictorially in \autoref{fig:struct_comp}.\footnote{Note that this functional representation does not explicitly show Batch Normalization's tunable parameters.}

\begin{equation} \label{eq:1}
f_{c}(x_i, W_{f_{c}}, b_{f_{c}}) \approx f_{2}(f_{1}(x_i, W_{f_1}, b_{f_1}), W_{f_2}, b_{f_2})  
\end{equation}

This new layer, $f_c$, given input $x_i$, contains learned weight matrix $W_{f_c}$ and bias vector $b_{f_c}$, is thus tasked with approximating the final representations learned by the two previously existent layers, $f_1$ and $f_2$. Where $f_2$ is fed the output of $f_1$, and has learned parameters $W_2$ and $b_2$. Likewise, $x_i$ is the input to $f_1$, and $f_1$ is parameterized by learned weight matrix $W_1$ and bias vector $b_1$. Given this compressed layer, we then test various initialization and training strategies in our deep net triage experiments.

\section{Experiments}

We applied six permutations of initialization and training compression experiments to the baseline models trained on the four chosen data sets. We conducted these experiments for each of the five layers blocks of VGG16 leading to a total of 124 networks. We chose to focus on the accuracy and the rate of convergence of these networks as measures of the criticality of the layer. Upon these findings, we further conducted qualitative filter activation analyses to verify the results.

Below, the initialization and training experiments are described, in addition to what each implies about the compressed model. This is followed by a brief nod to the data sets for those unfamiliar, and a description of the hyperparameters and framework used during optimization. 

\subsection{Initialization Methods}

\subsubsection{Random Initialization (RW)}
In this initialization scheme, the weights of the compressed layer are randomly initialized. More concretely they are initialized as samples from a Glorot Uniform Distribution \cite{glorot}. This \emph{tabula rasa}, could be thought of as allowing the compressed network to stray from the features already learned by the parent model and possibly find a more optimal set of filters.
\subsubsection{Mean Parent Initialization (MW)}
Via this scheme, an average of the weights in the parent's layer block compressed in the child model is used to initialize the compressed layer. The fact that we use a strict, unweighted average implies that while we don't know which of the parent's $N$ filters will be most useful, we know that filters similar to the parent's would best represent the data. Therefore, we load the average, $f_{avg}$, of the $i$ filters contained in the parent's block of uncompressed layers.

\begin {equation} \label{eq:2}
f_{avg} = \frac{1}{N} \sum_{i=1}^N{f_i} \quad where \quad f_i \in \mathbb{R}^{3x3x1}
\end {equation}

\subsubsection{Student-Teacher Network Initialization (STN)}
Like intermediate hints from \cite{fitnets} we use a Student-Teacher Network framework to teach the compressed network to give the same intermediate tensor output as the parent after the compressed layer. More concretely, we evaluate an L2 loss at the output of the compressed layer (after the max pooling layer), and at the output of the uncompressed block of layers in the parent network. We only allow for the gradients to update the compressed layer during this process, as it is used as a form of initialization. This initialization method can be thought of as using the parent to guide the optimization of the child. The implementation of the L2 loss function is given below.

\begin{equation} \label{eq:3}
\mathcal{L_{STN}} = \min_{W_s, b_s} \frac{1}{N} \sum_{i=1}^{N}{\| s(x_i, W_{s}, b_s) - t(x_i, W_{t}, b_t) \|^2}
\end{equation}

Here, $\mathcal{L_{STN}}$ is the loss function for our Student-Teacher Network Initialization. $N$ is the number of samples in a batch. And $x_i$ is the sample input to VGG16 models $s$ and $t$ with respective weight matrices $W_s$ and $W_t$.

\subsection{Training Methods}
\subsubsection{Frozen Model Weights Training (FM)}
In this training scheme, after the block of layers from the parent model has been compressed and initialized, we freeze all weights outside of the compressed layer. The notion behind this, is that a compressed layer can learn to work within the existing framework of the parent model, and also thus that the representations learned by the parent are most optimal.
\subsubsection{Thawed Model Weights Training (TM)}
Finally in this training strategy, we allow for gradient updates across the entire model after the layer has been compressed. This lends to the notion that the parent model's representations may not be optimal for the child network, or that even if they are, they must also be updated to optimally accommodate the new, compressed layer. 

\subsection{Data Sets}
The MNIST, CIFAR10, CIFAR100, and Stanford Dogs data sets were used to uncover generalizable trends across data sets \cite{mnist,cifar,stanford_dogs}. Below is a very brief description of the data sets to help reinforce this notion of varying complexity.
\paragraph{MNIST}
MNIST contains 60,000 training and 10,000 testing 28x28 greyscale images of digits from 10 classes.
\vspace*{-\baselineskip}
\paragraph{CIFAR10}
CIFAR10 contains 50,000 training and 10,000 testing 32x32 RGB images of objects and animals separated, in this case, into 10 classes.
\vspace*{-\baselineskip}
\paragraph{CIFAR100}
CIFAR100 contains 50,000 training and 10,000 testing 32x32 RGB images of objects and animals now separated into 100 classes.
\vspace*{-\baselineskip}
\paragraph{Stanford Dogs}
Stanford Dogs contains 12,000 training and 8,500 testing variously sized RGB images of dog breeds separated into 120 classes.

\subsection{Hyperparameters and Training}
We used the VGG16 network as first presented in \cite{vgg16} with Batch Normalization \cite{batch_norm}. We upsampled the inputs to (224,224,3) when necessary, zero-centered and normalized, and augmented the data with horizontal flips. Each baseline data set model was trained from scratch. All experiments across a given data set used the same hyperparameters. These are given in \autoref{table:ds}. Those which did not change in any instance are given in \autoref{table:const}. Keras' ReduceLROnPlateau was used to adaptively update the learning rate \cite{keras}.\footnote{Code to be released at time of publication.}

\begin{table}[t]
\begin{scriptsize}
\begin{center}
 \begin{tabular}{|c||c | c | c | c | c ||} 
 \hline
 \multicolumn{6}{|c|}{Data Set Specific Hyperparams.} \\
 \hline
 &LR$_1$ & LR$_{min}$ &  LR Decay & W. Decay & Epochs \\ [0.5ex] 
 \hline\hline
 MNIST & .001 & .00001 & .5 & .001 & 100 \\ 
 \hline
 CIFAR10 & .001 & .00001 & .7 & .0005 & 100 \\
 \hline
 CIFAR100 & .01 & .00001 & .7 & .0005 & 125 \\
 \hline
 Standford Dogs & .001 & .0000001 & .7 & 125 \\
 \hline 
\end{tabular}
\caption{\footnotesize{LR$_1$- initial learning rate, LR$_{min}$- minimum learning rate, LR Decay- decay fraction per plateau, W. Decay- Weight Decay per epoch, Epochs- number epochs base model trained}}
\label{table:ds}
\end{center}
\end{scriptsize}
\end{table}

\begin{table}[t]
\begin{scriptsize}
\begin{center} 
 \begin{tabular}{|c||c | c | c | c | c ||} 
 \hline
 \multicolumn{6}{|c|}{Constant Params.} \\
 \hline
 & $\rho$ & Pat. &  CD & Comp. & STN \\ [0.5ex] 
 \hline\hline
 ALL & .9 & 1 & 3 & 25 & 12 \\ 
 \hline
\end{tabular}
\caption{$\rho$- Momentum, Pat.- patience of learning rate plateau, CD- cooldown of learning rate plateau, Comp.- epochs trained after structural compression, STN- epochs STN was trained for}
\label{table:const}
\end{center}
\end{scriptsize}
\end{table}

\begin{figure*}[!ht]
\begin{tabular}{cc}
\centering
\includegraphics[width=.45\linewidth]{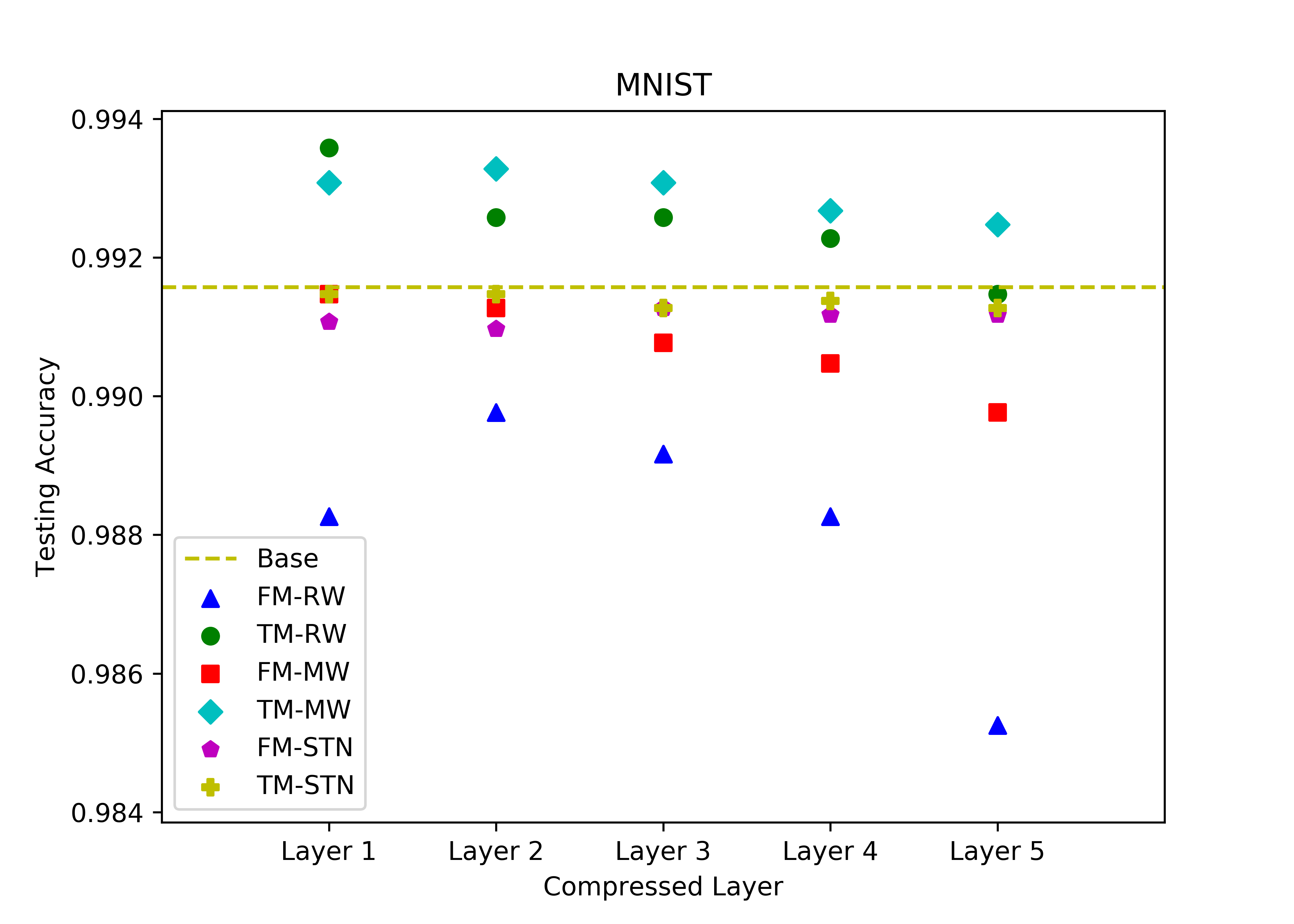} &
\includegraphics[width=.45\linewidth]{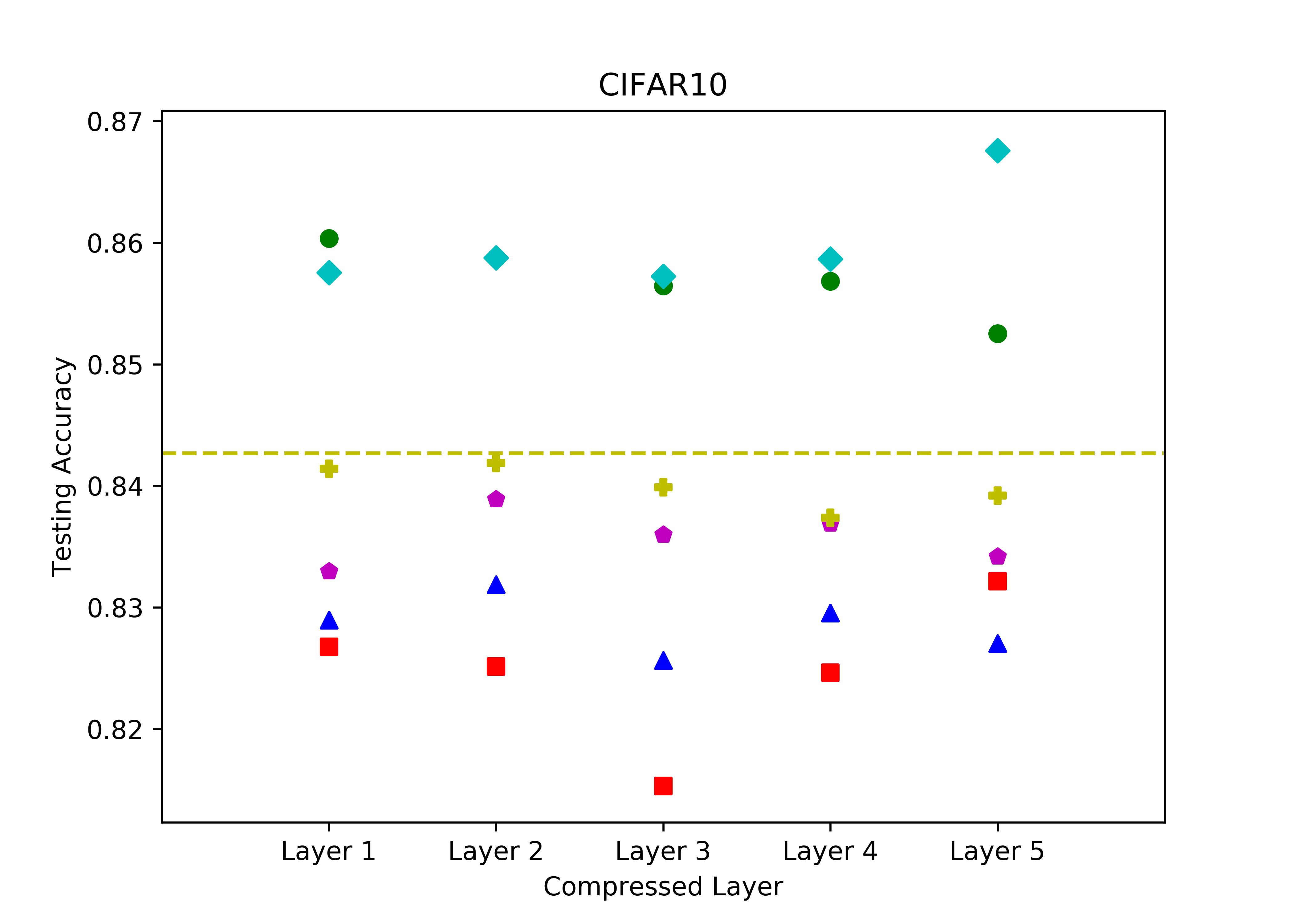} \\
\includegraphics[width=.45\linewidth]{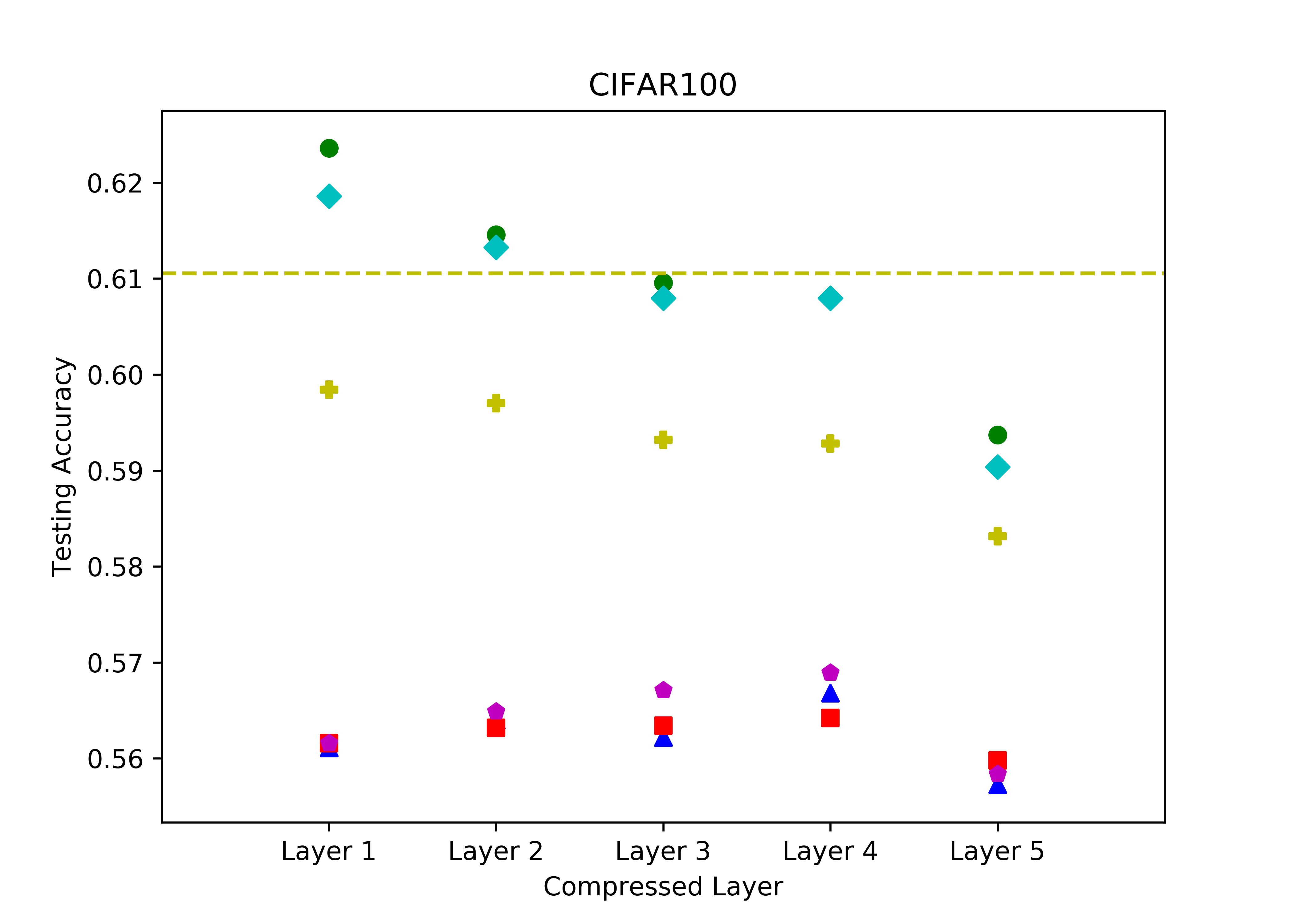} &
\includegraphics[width=.45\linewidth]{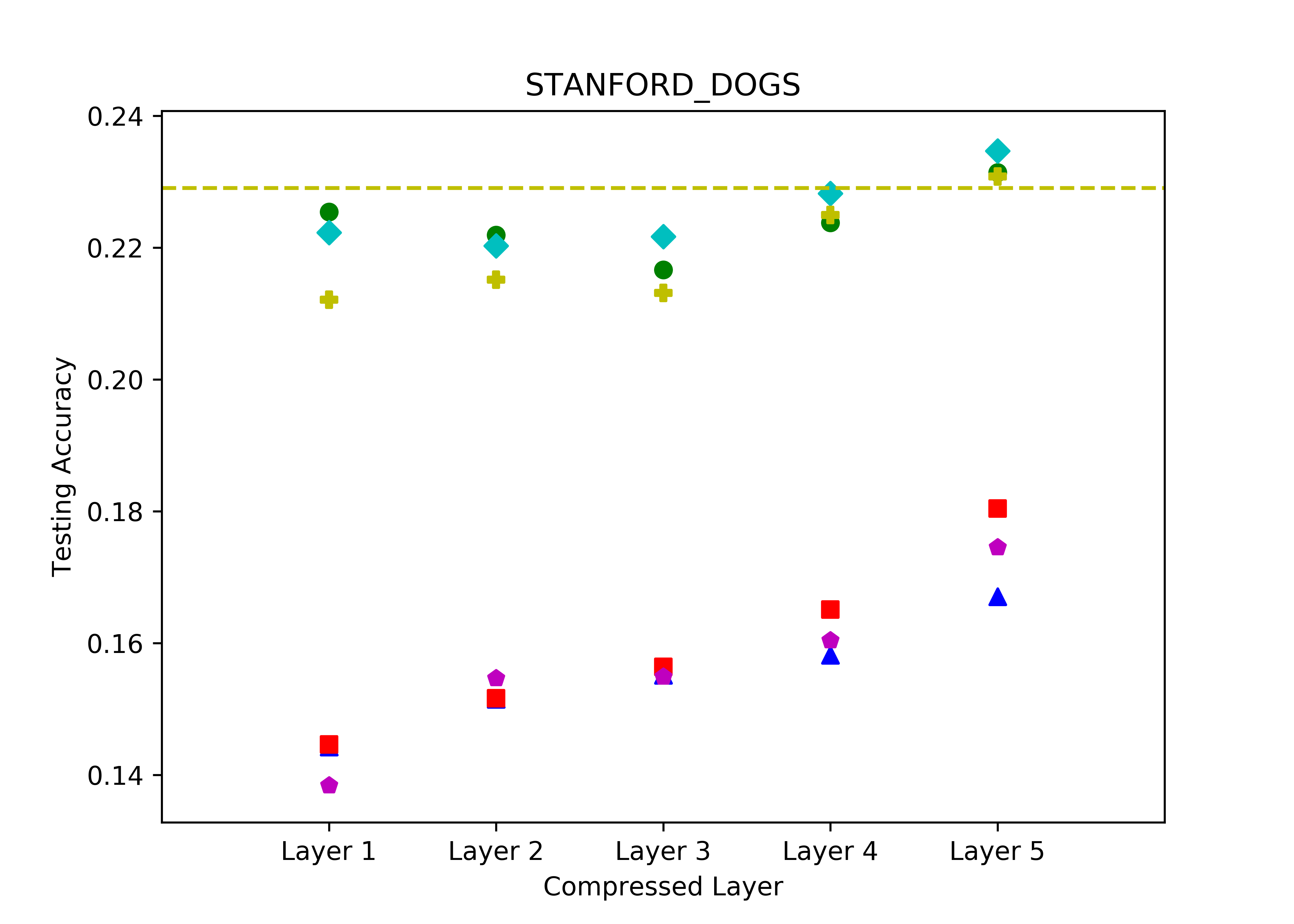} \\
\end{tabular}
\caption{For each data set, experiments are given in terms of which layer was compressed. The six tested experimental permutations are shown along with the baseline achieved by the uncompressed network.}
\label{fig:max_acc}
\end{figure*}

\section{Analysis and Results}

We conducted every permutation of the three initialization and two training strategies after structural compression for every layer block and each data set. We first looked at the maximum accuracy achieved by each. We then assessed the rate at which each model converged to these accuracies.

In \autoref{fig:max_acc} we show the maximum accuracies attained by each model. We first note that no layer is more critical than any other. This is of surprise because as later layers are compressed, more parameters, and furthermore, higher-level representations, are removed from the model.

Upon further inspection of \autoref{fig:max_acc} another trend is immediately evident: that FM models---those where only the compressed layer was allowed to train after compression---achieve significantly lower relative accuracy. This refutes the idea that a layer injected into a model can learn to fit into the representations of the existing model. One may also note that TM models---those that were fine-tuned across the whole model---often out perform the baseline model. We believe this to be a factor of overfitting, as found in \cite{visandunderstand}. We lastly note that TM-STN seems to break this trend. At first on the MNIST data set, TM-STN tends to perform poorly and on par with FM-STN. Then as the data set complexity increases, its relative performance increases to that of the other FM methods on Stanford Dogs. Though it never outperforms the baseline. This pattern encouraged us to visualize the filter activations for this method.

\begin{figure*}[ht]
\centering
\includegraphics[width=0.99\linewidth]{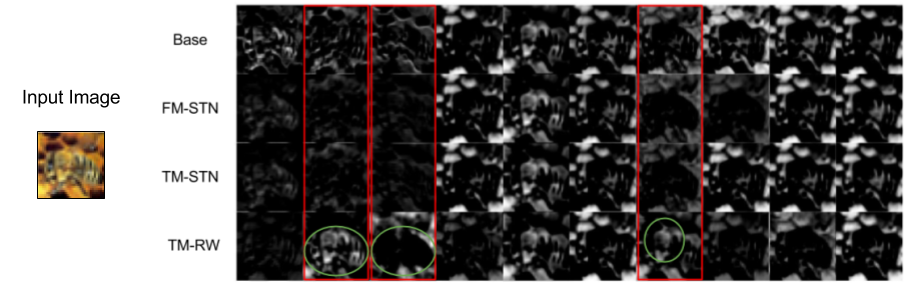}
\caption{Activations at the output of Conv. Block 1 from the Base, FM-STN, TM-STN, and TM-RW methods. Red boxes denote the filters which contain similar Base, FM-STN, and TM-STN filters, but differing TM-RW filters. Green circles help identify those differences for the reader.}
\label{fig:act}
\end{figure*}

\begin{figure*}[b]
\centering
\begin{tabular}{cc}
\includegraphics[width=.45\linewidth]{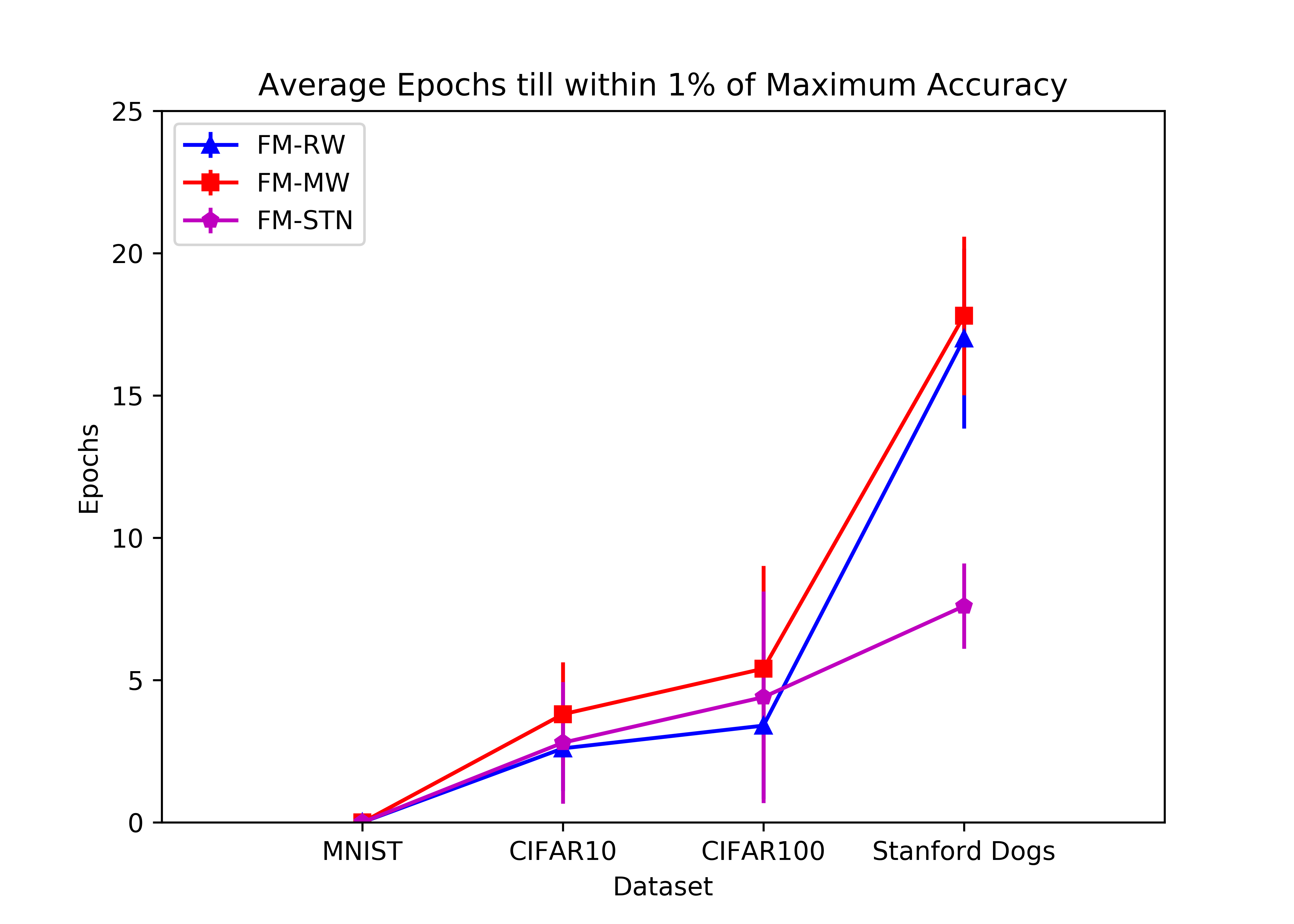} &
\includegraphics[width=.45\linewidth]{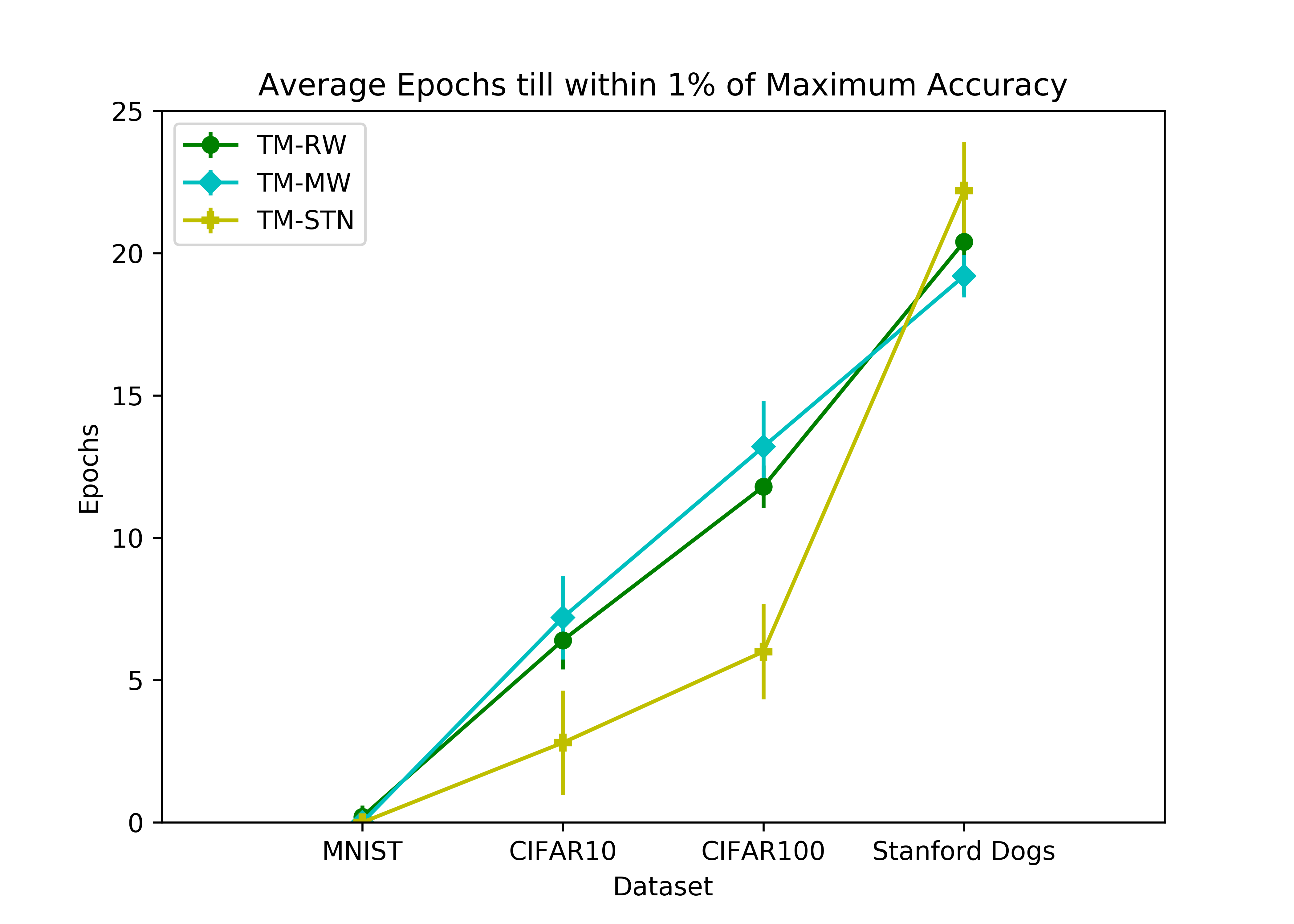} \\
\end{tabular}
\caption{The average number of epochs for each compressed model to converge to within 99\% of maximal accuracy.}
\label{fig:convergence}
\end{figure*}

Specifically, we chose to visualize the filters of four variants of the CIFAR10 model: baseline, FM-STN, TM-STN, and TM-RW. We chose this data set and these models because in \autoref{fig:max_acc} it is clear that on CIFAR10 the difference between these methods were the most distinct. We can see that even after the compression of Layer 1, that FM-STN and TM-STN are distinctly separated, that TM-STN is very close to baseline, and that accuracy on TM-RW is significantly higher than that of TM-STN and the baseline. This implies that any differences in filters between these methods might be most distinct on the CIFAR10 data set.  

We used an image chosen at random from the data set and visualized the filters at the output of the ReLU activation layer after the compressed layer for the compressed models, and after Conv 2 of Block 1 on the baseline model. We wanted to assess what features Block 1 was responsible for learning in each case. In \autoref{fig:act} we show the first ten filters of these activations. Upon close comparison, one can see that the activation responses learned for the base and two STN networks are similar if not identical. While the STN filters are often, less focused and less bright, the filter's response to the features in the image are nearly identical. This is not the case for the filters learned by TM-RW. In the green circles, we can see sources of activation that differ from the other three models.

These observations indicate that the STN networks have been forced to learn the same filter responses as the parent network. Meanwhile, the TM-RW network has learned novel filters, differing from the parent network, which are more optimally representative. This seems to show why a compressed network can outperform its parent: because it has found a more optimal set of features. This is additionally reflected by the observation that in \autoref{fig:max_acc} no STN network outperforms the baseline network. 

We now turn our focus to the rate of convergence of these compressed networks. In \autoref{fig:convergence} there are two plots---one for FM networks, and the other for TM networks. These show how quickly each network converges to within 1\% of its maximal accuracy. This indicates not at which epoch maximal accuracy was achieved, but rather how long it took for the network to converge. We can see that as data set complexity increases, so too does the time it takes for each network to converge. Additionally, though, we can see two things. Firstly, that initialization with an average set of weights from the parent network not only doesn't help, it may even reduce the rate of convergence. Secondly, the STN models both appear to converge more quickly than the other models, but on differing data sets. While the FM-STN converged significantly faster than the other models on Stanford Dogs, the TM-STN converged even later. Similarly, the TM-STN converged more quickly on CIFAR10 and CIFAR100, but the FM-STN didn't.

This may be because on the simpler data sets, every model is able to over fit thereby allowing the STN and FM models to converge and converge quickly. Meanwhile, on Stanford Dogs, no model is able to do well and no overfitting occurs. Therefore the TM-STN accuracy can improve to the performance of the other TM models, and the FM-STN, at a lower accuracy, can continue to converge more quickly then its fellow FM models. This seems to indicate that generally STN networks can help increase the rate of convergence of a model.

\section{Conclusions and Future Work}

We present a novel method for analyzing deep neural networks which we refer to as \emph{deep net triage}. By drawing experiments from existing network compression and network analysis methods, such as Knowledge-Distillation and filter activation visualization, we sequentially apply \emph{structural compression} to compress sections of a parent network which then allow us to assess the \emph{criticality} of each layer individually.

We show through our analysis that no layer is more critical than another. We additionally show, that rather than a layer being able to optimize to an existing network, the entire network must be allowed to fine-tune in order fully integrate a new layer. Furthermore, that when a compressed model is allowed to fine-tune, it is able to learn more representative weights which can lead to increased performance. We lastly show that while Student-Teacher Networks can improve the rate of convergence of a large network model, it cannot increase the student performance beyond that of its teacher. These findings help build intuition and further understanding of how these layers perform and how they can be improved. 

In the future, we hope to build upon the structure of this analysis to further probe the workings of neural networks. While not statistically significant, we noted that the accuracies for each data set seem to qualitatively follow a pattern that is unique to the data set. Additionally, to assess the effect of overfitting on this exercise, we'd like to compress models structurally to the point at which they fail. This combined with the compression method from \cite{dally-pruningtraininghuffman} could yield close to optimally compressed network models. Lastly, further analysis of more thoughtful initialization methods and on other, more modern neural network models could be insightful.

Through this work, we hope to provide a structured way to sequentially analyze individual portions of a neural network. We do so to gain a better intuition for the mechanisms at play within. While, as a community, we may continue to develop ever better performing methods for given problem spaces, we will never truly advance as a field until further intuition for and understanding of deep networks is developed. As theories are developed on one end, so too must experimental intuition be developed on the other.  

\bibliographystyle{named}
\bibliography{deep_net_triage}

\begin{thebibliography}{}

\bibitem[\protect\citeauthoryear{Bucila \bgroup \em et al.\egroup
  }{2006}]{caruana-modelcomp}
Cristian Bucila, Rich Caruana, and Alexandru Niculescu-Mizil.
\newblock Model compression.
\newblock 2006.

\bibitem[\protect\citeauthoryear{Chollet and others}{2015}]{keras}
Fran\c{c}ois Chollet et~al.
\newblock Keras.
\newblock \url{https://github.com/keras-team/keras}, 2015.

\bibitem[\protect\citeauthoryear{Glorot and Bengio}{2010}]{glorot}
Xavier Glorot and Yoshua Bengio.
\newblock Understanding the difficulty of training deep feedforward neural
  networks.
\newblock In Yee~Whye Teh and Mike Titterington, editors, {\em Proceedings of
  the Thirteenth International Conference on Artificial Intelligence and
  Statistics}, volume~9 of {\em Proceedings of Machine Learning Research},
  pages 249--256, Chia Laguna Resort, Sardinia, Italy, 13--15 May 2010. PMLR.

\bibitem[\protect\citeauthoryear{Haeffele and Vidal}{2015}]{vidal}
Benjamin~D. Haeffele and Ren{\'{e}} Vidal.
\newblock Global optimality in tensor factorization, deep learning, and beyond.
\newblock {\em CoRR}, abs/1506.07540, 2015.

\bibitem[\protect\citeauthoryear{Han \bgroup \em et al.\egroup
  }{2015}]{dally-pruningtraininghuffman}
Song Han, Huizi Mao, and William~J. Dally.
\newblock Deep compression: Compressing deep neural network with pruning,
  trained quantization and huffman coding.
\newblock {\em CoRR}, abs/1510.00149, 2015.

\bibitem[\protect\citeauthoryear{He \bgroup \em et al.\egroup }{2015}]{resnet}
Kaiming He, Xiangyu Zhang, Shaoqing Ren, and Jian Sun.
\newblock Deep residual learning for image recognition.
\newblock {\em CoRR}, abs/1512.03385, 2015.

\bibitem[\protect\citeauthoryear{Hinton \bgroup \em et al.\egroup
  }{2015}]{knowledge_distillation}
Geoffrey Hinton, Oriol Vinyals, and Jeff Dean.
\newblock Distilling the knowledge in a neural network.
\newblock {\em arXiv preprint arXiv:1503.02531}, 2015.

\bibitem[\protect\citeauthoryear{Ioffe and Szegedy}{2015}]{batch_norm}
Sergey Ioffe and Christian Szegedy.
\newblock Batch normalization: Accelerating deep network training by reducing
  internal covariate shift.
\newblock {\em CoRR}, abs/1502.03167, 2015.

\bibitem[\protect\citeauthoryear{Khosla \bgroup \em et al.\egroup
  }{2011}]{stanford_dogs}
Aditya Khosla, Nityananda Jayadevaprakash, Bangpeng Yao, and Li~Fei-Fei.
\newblock Novel dataset for fine-grained image categorization.
\newblock In {\em First Workshop on Fine-Grained Visual Categorization, IEEE
  Conference on Computer Vision and Pattern Recognition}, Colorado Springs, CO,
  June 2011.

\bibitem[\protect\citeauthoryear{Krizhevsky \bgroup \em et al.\egroup
  }{}]{cifar}
Alex Krizhevsky, Vinod Nair, and Geoffrey Hinton.
\newblock Cifar-10 (canadian institute for advanced research).

\bibitem[\protect\citeauthoryear{LeCun and Cortes}{2010}]{mnist}
Yann LeCun and Corinna Cortes.
\newblock {MNIST} handwritten digit database.
\newblock 2010.

\bibitem[\protect\citeauthoryear{LeCun \bgroup \em et al.\egroup }{1990}]{OBD}
Yann LeCun, John~S. Denker, and Sara~A. Solla.
\newblock Optimal brain damage.
\newblock In D.~S. Touretzky, editor, {\em Advances in Neural Information
  Processing Systems 2}, pages 598--605. Morgan-Kaufmann, 1990.

\bibitem[\protect\citeauthoryear{Mordvintsev \bgroup \em et al.\egroup
  }{}]{deepdream}
Alexander Mordvintsev, Christopher Olah, and Mike Tyka.
\newblock Inceptionism: Going deeper into neural networks.

\bibitem[\protect\citeauthoryear{Romero \bgroup \em et al.\egroup
  }{2014}]{fitnets}
Adriana Romero, Nicolas Ballas, Samira~Ebrahimi Kahou, Antoine Chassang, Carlo
  Gatta, and Yoshua Bengio.
\newblock Fitnets: Hints for thin deep nets.
\newblock {\em CoRR}, abs/1412.6550, 2014.

\bibitem[\protect\citeauthoryear{Simonyan and Zisserman}{2014}]{vgg16}
Karen Simonyan and Andrew Zisserman.
\newblock Very deep convolutional networks for large-scale image recognition.
\newblock {\em CoRR}, abs/1409.1556, 2014.

\bibitem[\protect\citeauthoryear{Suganuma \bgroup \em et al.\egroup
  }{2017}]{genetic}
Masanori Suganuma, Shinichi Shirakawa, and Tomoharu Nagao.
\newblock A genetic programming approach to designing convolutional neural
  network architectures.
\newblock In {\em Proceedings of the Genetic and Evolutionary Computation
  Conference}, pages 497--504. ACM, 2017.

\bibitem[\protect\citeauthoryear{Yosinski \bgroup \em et al.\egroup
  }{2015}]{deepvis}
Jason Yosinski, Jeff Clune, Anh Nguyen, Thomas Fuchs, and Hod Lipson.
\newblock Understanding neural networks through deep visualization.
\newblock {\em arXiv preprint arXiv:1506.06579}, 2015.

\bibitem[\protect\citeauthoryear{Zeiler and Fergus}{2013}]{visandunderstand}
Matthew~D. Zeiler and Rob Fergus.
\newblock Visualizing and understanding convolutional networks.
\newblock {\em CoRR}, abs/1311.2901, 2013.

\end{thebibliography}

\end{document}